\documentclass[10pt]{article} 
\usepackage[preprint]{rlc}

\usepackage{amssymb}            
\usepackage{mathtools}          
\usepackage{mathrsfs}           
\mathtoolsset{showonlyrefs}     
\usepackage{graphicx}           
\usepackage{subcaption}         
\usepackage[space]{grffile}     
\usepackage{url}                

\usepackage{wrapfig}
\usepackage{hyperref}
\usepackage{booktabs}
\usepackage{xspace}
\usepackage{pifont}
\usepackage{xcolor}
\usepackage{multirow}
\usepackage{colortbl}
\usepackage{tabularx}
\usepackage{enumitem}
\usepackage{adjustbox}
\usepackage{ragged2e}   
\usepackage{makecell} 
\usepackage{array}       
\usepackage{tcolorbox}
\usepackage{listings}
\usepackage{tikz}
\newtcolorbox[auto counter]{response}[1][]{
    top=10pt,
    bottom=10pt,
    left=15pt,
    right=15pt,
    colback=gray!5,
    colframe=black,
    colbacktitle=black!80,
    fonttitle=\bfseries,
    coltitle=white,
    arc=1mm,              
    title=Prompt~\thetcbcounter: #1
    fonttitle=\bfseries\normalsize,        
    fontupper=\small,             
    width=\linewidth,                  
}

\newcommand{\ie}{\emph{i.e.}}

\usepackage{tocloft}
\usepackage{tikz}

\newcommand{\gradientColorBench}{%
    \textcolor{red!10!orange}{C}%
    \textcolor{red!50!orange}{o}%
    \textcolor{magenta!70!red}{l}%
    \textcolor{purple!80!magenta}{o}%
    \textcolor{purple!40!violet!70!magenta}{r}%
    \textcolor{purple!30!violet!70!magenta!80!blue}{B}%
    \textcolor{purple!60!magenta!50!blue}{e}%
    \textcolor{purple!60!magenta!35!blue}{n}%
    \textcolor{purple!60!magenta!30!blue!80!cyan}{c}%
    \textcolor{purple!60!magenta!30!blue!60!cyan}{h}%
}

\title{\gradientColorBench: Benchmarking Mobile Agents with Graph-Structured Framework for Complex Long-Horizon Tasks}

\author{\\
\name{Yuanyi Song}$^{1}$\thanks{This work was done during Yuanyi Song’s internship at OPPO.}\textsuperscript{~ †},
\name{Heyuan Huang}$^{2}$\thanks{Y. Song and H. Huang contributed equally to this research.},
\name{Qiqiang Lin}$^{2}$,
\name{Yin Zhao}$^{2}$,
\name{Xiangmou Qu}$^{2}$, \\
\name{Jun Wang}$^{2}$,
\name{Xingyu Lou}$^{2}$\textsuperscript{‡},
\name{Weiwen Liu}$^{1}$,
\name{Zhuosheng Zhang}$^{1}$, 
\name{Jun Wang}$^{2}$,\\
\name{Yong Yu}$^{1}$,
\name{Weinan Zhang}$^{1}$\textsuperscript{‡},
\name{Zhaoxiang Wang}$^{2}$\thanks{X. Lou, W. Zhang and Z. Wang are the corresponding authors.} \vspace{+0.1cm}\\ 
$^1$Shanghai Jiao Tong University \quad
$^2$OPPO\\
\texttt{norsheep919@sjtu.edu.cn}\quad \texttt{louxingyu@oppo.com}\\
\texttt{wnzhang@sjtu.edu.cn}\quad \texttt{steven.wangzx@gmail.com}
}

\begin{document}

\maketitle

\begin{abstract}
The rapid advancement of multimodal large language models has enabled agents to operate mobile devices by directly interacting with graphical user interfaces, opening new possibilities for mobile automation. However, real-world mobile tasks are often complex and allow for multiple valid solutions. This contradicts current mobile agent evaluation standards: offline static benchmarks can only validate a single predefined ``golden path'', while online dynamic testing is constrained by the complexity and non-reproducibility of real devices, making both approaches inadequate for comprehensively assessing agent capabilities. To bridge the gap between offline and online evaluation and enhance testing stability, this paper introduces a novel graph-structured benchmarking framework. By modeling the finite states observed during real-device interactions, it achieves static simulation of dynamic behaviors. Building on this, we develop ColorBench, a benchmark focused on \underline{\textbf{co}}mplex \underline{\textbf{lo}}ng-ho\underline{\textbf{r}}izon tasks. It supports evaluation of multiple valid solutions, subtask completion rate statistics, and atomic-level capability analysis. ColorBench contains 175 tasks (74 single-app, 101 cross-app) with an average length of over 13 steps. Each task includes at least two correct paths and several typical error paths, enabling quasi-dynamic interaction. By evaluating ColorBench across various baselines, we discover limitations of existing models and propose improvement directions and feasible technical pathways to enhance agents' performance on complex, long-horizon problems based on experimental results. Code and data are available at: \href{https://github.com/MadeAgents/ColorBench}{ColorBench}.
\end{abstract}

\section{Introduction}


As one of the primary human-computer interaction entry points for today's internet, mobile devices present an urgent need to enhance internet service accessibility and user experience by exploring their automation, calling for AI agents' graphical user interface (GUI) interaction capabilities~\citep{wen2024autodroid, hong2024cogagent,li2025autogui,chen2025less}.
With the advancement of artificial intelligence, utilizing multimodal large language models (MLLMs) as agents to operate graphical user interfaces GUIs for mobile tasks has gained increasing attention~\citep{jiang2025appagentx,cheng2025kairos,wang2024mobile,wang2024mobile2,wu2025quick}, giving rise to numerous outstanding benchmarks for mobile tasks~\citep{rawles2023androidinthewild,lu2024gui,sun2022meta,li2024effects,liu2025learnact,huang2025mvisu}.

Existing mobile GUI agent benchmarks can be broadly categorized into two paradigms~\citep{zhang2024large,xu2025mobile}: 

\begin{wrapfigure}{r}{0.50\textwidth}
  \centering
  \includegraphics[width=0.9\linewidth]{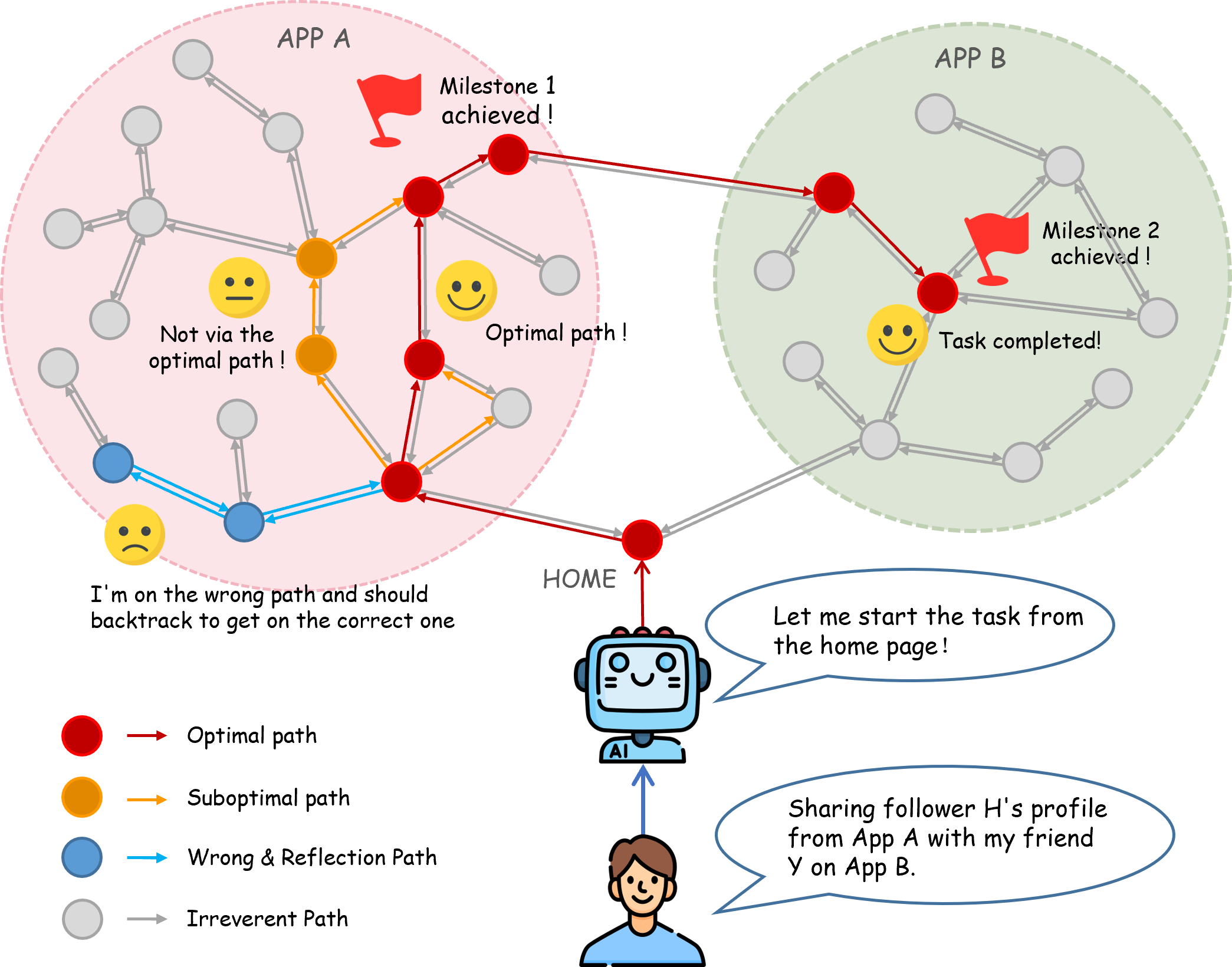}
  \caption{\textbf{Structure of the ColorBench.} \textmd{It illustrates multi-path solutions, reflective backtracking and automated evaluation milestones, demonstrating how the graph is structured and utilized. Each node represents a screen list.}}
  \label{fig:graph_demo}
  \vspace{-6pt}
\end{wrapfigure}

\textbf{1) Offline Static Evaluation:} 
This paradigm involves assessing agent performance using static image trajectory data. Although widely adopted due to the feasibility of rapid, large-scale data collection, this method is subject to several prominent limitations: 
a) \textit{Rigid Assessment.} It relies on fixed trajectories for evaluation, failing to assess multiple potential solutions and leading to potential misjudgments.
b) \textit{Oversimplified Metrics.} Its evaluation metrics are singular, overemphasizing step-level success rates while lacking a comprehensive assessment of overall task completion.
c) \textit{Coarse-Grained Diagnosis.} Its evaluation dimensions are broad, considering only task-wise completion status and lacking fine-grained analysis of atomic capabilities.
These issues cause a \textbf{significant ``offline-online evaluation discrepancy'' during static testing}, which means an agent's performance on offline static evaluations may not positively correlate with its actual device performance~\citep{li2024effects,sun2022meta}.

\textbf{2) Online Dynamic Evaluation:} This paradigm assesses models within a dynamic environment to deliver more authentic performance metrics. It can be further categorized into two types: virtual environment evaluation and real-device evaluation. However, existing virtual testing environments like AndroidWorld~\citep{rawles2024androidworld} and AndroidLab~\citep{xu2024androidlab} involve applications and interaction patterns that differ from real-world scenarios, making it challenging to accurately reflect an agent's true deployment performance. Moreover, its implementation is also relatively challenging. While real-device evaluation is highly valued for its ability to accurately reflect user scenarios, it suffers from the following defects: a) \textit{Instability}. Variations in page loading delays and sudden ad pop-ups may cause unexpected interruptions, leading to ambiguous evaluation criteria; 
b) \textit{Inefficient Result Assessment.} The highly dynamic GUI makes automated checks difficult, forcing reliance on time-consuming manual verification~\citep{dai2025advancing,hu2024auitestagent}. 
c) \textit{Security Risks.} Most applications require account login to access full functionality, posing risks of unintended payments, data deletion, or personal information leaks during evaluations~\citep{ma2024caution,chen2025harmonyguard,zhang2024psysafe}. These factors directly result in \textbf{poor reproducibility and low efficiency during dynamic testing}.

To address these problems, we propose \textbf{ColorBench}, a novel \textbf{graph-structured} mobile agent \textbf{benchmark for \underline{co}mplex \underline{lo}ng-ho\underline{r}izon tasks.} 
As presented in Table 1, ColorBench aims to strike a balance between offline static and online dynamic evaluation through a finite-state simulation, maintaining the stability of the former while incorporating the flexibility of the latter. Moreover, we focus on complex long-horizon tasks due to their composite nature, which is characterized by multiple atomic subtasks executed in sequential, parallel, or recursive patterns, along with the existence of multiple path solutions, thereby making a graph-structured benchmark ideally suited for this context.

\begin{table}[t]
    \centering
    \fontsize{8.5pt}{9.8pt}\selectfont
    \renewcommand\arraystretch{1.05}
    \caption{\textbf{Comparison between ColorBench and other mobile agents benchmarks.} \textmd{Our ColorBench is a static graph environment but supports interaction similar to real world. The column ``Interaction'' indicates whether to support the agent to interact with provided environment when evaluating. The column ``AtomCap'' means whether to support atomic capability assessment. The ``Step'' only calculates the optimal solution for each task and does not account for randomly appearing advertisements or longer correct paths. It is worth mentioning that the benchmark statistics in the third section include the training set.}}
    \vspace{-2mm}
    \begin{tabular}{lcccccc}
    \toprule
    \textbf{\makecell[l]{Dataset}} & \textbf{\makecell{\# Tasks}} & \textbf{\makecell{\# Step}} & \textbf{\makecell{\# Apps}} & \textbf{\makecell{Inter-\\action}} & \textbf{\makecell{Multiple\\Solution}} & \textbf{\makecell{Atom-\\Cap}}  \\
    \midrule
    AndroidLab~\citep{xu2024androidlab} & 138 & 8.5 & 9 & \textcolor{green!80!black}{\ding{51}} & \textcolor{green!80!black}{\ding{51}} & \textcolor{red}{\ding{55}}\\
    AndroidWorld~\citep{rawles2024androidworld} & 116 & - & 20 & \textcolor{green!80!black}{\ding{51}} & \textcolor{green!80!black}{\ding{51}} & \textcolor{red}{\ding{55}}\\
    Mobile-Env~\citep{zhang2023mobile} & 150 & - & -  & \textcolor{green!80!black}{\ding{51}} & \textcolor{green!80!black}{\ding{51}} & \textcolor{red}{\ding{55}}\\
    \midrule
    MobileAgentBench ~\citep{wang2024mobileagentbench} & 100 & - & 10 & \textcolor{red}{\ding{55}} & \textcolor{green!80!black}{\ding{51}} & \textcolor{red}{\ding{55}}\\
    UI-NEXUS~\citep{guo2025atomic} & 100 & 14.05 & 50 & \textcolor{red}{\ding{55}} & \textcolor{green!80!black}{\ding{51}} & \textcolor{red}{\ding{55}}\\
    Mobile-Eval~\citep{wang2024mobile} & 33 & 5.5 & 10  & \textcolor{red}{\ding{55}} & \textcolor{green!80!black}{\ding{51}} & \textcolor{red}{\ding{55}} \\
    Mobile-Eval-E~\citep{wang2025mobile} & 25 & 14.56 & 15  & \textcolor{red}{\ding{55}} & \textcolor{green!80!black}{\ding{51}} & \textcolor{red}{\ding{55}} \\
    SPA-BENCH~\citep{chen2024spa} & 340 & 8.2 & 66 & \textcolor{red}{\ding{55}} & \textcolor{green!80!black}{\ding{51}} & \textcolor{red}{\ding{55}} \\
    MVISU-Bench~\citep{huang2025mvisu} & 404 & - & 137 & \textcolor{red}{\ding{55}} & \textcolor{green!80!black}{\ding{51}} & \textcolor{red}{\ding{55}}\\
    AppAgent~\citep{zhang2025appagent} & 50 & - & 10  & \textcolor{red}{\ding{55}} & \textcolor{green!80!black}{\ding{51}} & \textcolor{red}{\ding{55}} \\
    \midrule
    AndroidControl~\citep{li2024effects} & 15283 & 4.8 & 833 & \textcolor{red}{\ding{55}} &  \textcolor{red}{\ding{55}} & \textcolor{red}{\ding{55}}\\
    GUI-Odyssey~\citep{lu2024gui} & 7735 & 15.4 & 201  & \textcolor{red}{\ding{55}} &  \textcolor{red}{\ding{55}} & \textcolor{red}{\ding{55}}\\
    Meta-GUI~\citep{sun2022meta} & 1125 & 4.3 & -  & \textcolor{red}{\ding{55}} &  \textcolor{red}{\ding{55}} & \textcolor{red}{\ding{55}}\\
    Mobile-Bench-v2~\citep{xu2025mobile} & 12,856 & 7.28 & 49 & \textcolor{red}{\ding{55}} & \textcolor{green!80!black}{\ding{51}} & \textcolor{red}{\ding{55}}\\
    \midrule
    \textbf{ColorBench} & 175 & 13.13$+$ & 21 & \textcolor{green!80!black}{\ding{51}} & \textbf{\textcolor{green!80!black}{\ding{51}}} &\textcolor{green!80!black}{\ding{51}} \\
    \bottomrule
    \end{tabular}
    \vspace{-2em}
    \label{tab:datasets}
\end{table}

We organize the colorbench into a strongly connected graph as shown in Figure~\ref{fig:graph_demo}, where mobile screen states serve as nodes and action transition relationships between nodes serve as edges,  multiple solutions (paths of different colors), reflective backtracking (the blue path), and automated evaluation milestones (the red flag). Specifically, we designed an \textbf{efficient graph-structured benchmark construction methodology}, enabling subsequent researchers to expand
or reconstruct graphs. It collects static trajectories via breadth-first and depth-first algorithms, merges nodes and edges by computing screenshot similarity, and automatically labels bounding boxes as transition conditions. Manual validation is integrated throughout to ensure data quality.

\begin{figure}[htb]
  \centering
  \includegraphics[width=\textwidth]{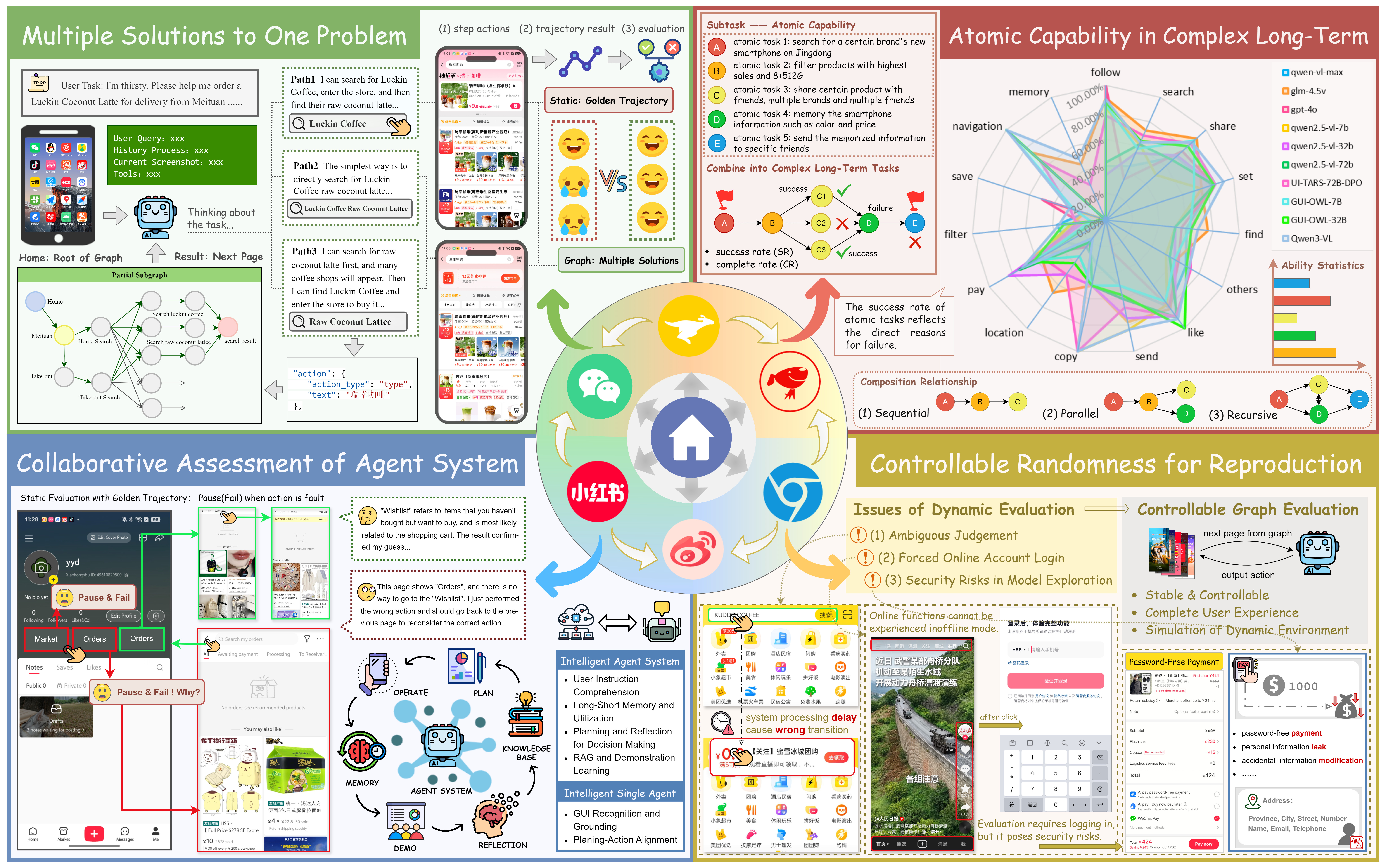}
  \caption{\textbf{Advantage of graph-structured benchmark.} \textmd{Our constructed ColorBench possesses these advantages. The upper-left corner demonstrates multiple solutions to one problem, supporting diverse execution paths for the same task. The lower-left corner illustrates how the graph's strongly connected structure effectively enables model collaboration capabilities such as backtracking and reflection. The upper-right corner showcases the graph's inherent suitability for complex long-horizon tasks and its support for atomic capability evaluation. The lower-right corner highlights the graph's superior controllability and reproducibility compared to dynamic evaluation.}}
  \label{fig:main-image}
  \vspace{-4mm}
\end{figure}

The graph-structured benchmark effectively resolves the aforementioned limitations present in the existing benchmarks. As shown in Figure~\ref{fig:main-image}, it mainly has the following key advantages:
\textbf{1) Multi-Solution Evaluation:} It supports the evaluation of multiple valid solutions for a single task;
\textbf{2) Enhanced Collaboration:} It enables agents to fully utilize collaborative abilities such as reflection and backtracking, thereby preventing the underestimation of model capabilities due to environmental constraints;
\textbf{3) Atomic Assessment}: By setting milestones for subtasks, it enables stable and automated evaluation as well as fine-grained assessment of \textit{atomic task capabilities} for diagnosing atomic-task level weaknesses that lead to agent failures;
\textbf{4) Controllable Environment}: It provides a secure, stable, and controllable testing environment that remains statically reliable while supporting rich interactions.
Together, these properties allow the framework to effectively bridge the gap between offline and online evaluation.

We carefully select three commonly used closed-source models and ten open-source models, alongside two customized baselines, for a comprehensive evaluation on ColorBench. Through extensive experiments, we systematically reveal the limitations of existing agents in tackling complex long-horizon tasks and diagnose their underlying causes. Based on these findings, we provide concrete recommendations for developing more capable agents suitable for such challenging scenarios.

Overall, the primary contributions of our work can be summarized as follows:
\begin{itemize}
\item \textbf{Graph-Structured Benchmark.} We propose a benchmark framework of graph structure that bridges the discrepancy between offline and online tests
We design an effective construction methodology that balances quality and efficiency and verify the significance and feasibility of the graph paradigm by statistical experiments.
\item \textbf{ColorBench for Complex Long-Horizon Task.} It is the first comprehensive graph-structured mobile agent benchmark for complex long-horizon tasks. By extending step-level evaluation to the atomic-task level, we assess and pinpoint models' weaknesses in atomic capabilities through evaluations within complete complex tasks.
\item \textbf{Pathways and Insights for ColorBench Solutions.}  
We conduct systematic experiments and analyses on ColorBench and offer improvement directions and insights for future complex long-horizon task solutions based on experimental findings.
\end{itemize}

\section{Related Work}
\subsection{Mobile GUI Agent Benchmark} 

Traditional GUI agent evaluation methods can be broadly categorized into two types: offline static evaluation based on trajectory chains, and end-to-end online dynamic evaluation~\citep{rawles2023androidinthewild,sun2022meta,li2024effects}. Dynamic evaluation can be further subdivided into virtual sandbox environments and real devices~\citep{rawles2024androidworld, xu2024androidlab,zhang2023mobile}. For applications requiring login credentials, there is no fundamental difference in their impact on the real world. Static evaluation is straightforward and comprehensive. Using image-answer pairs alone, benchmarks can be created for step tasks such as UI recognition, grounding capabilities\citep{deka2017rico, wang2021screen2words, li2020widget},
page transition relationship identification~\citep{chen2024guicourse}, and simple atomic tasks~\citep{li2024effects,deng2023mind2web,taleby2016innovative,zhang2024android,wu2025quick},
as well as demonstration learning capabilities~\citep{liu2025learnact}. However, for complex long-horizon tasks composed of multiple atomic tasks or even spanning multiple apps, multiple solutions often exist~\citep{guo2025atomic,lu2024gui,wang2025mobile}. Static datasets, constrained by their single standard answer, struggle to accurately assess real-world model capabilities~\citep{wang2025odysseybench}. 
Existing GUI benchmarks for complex long-horizon tasks across various platforms and scenarios typically employ dynamic testing in real or simulated environments~\citep{liu2025verigui,ye2025realwebassist}.
Unlike the web environment, mobile GUI operations lack accessible URLs, and inherent issues with mobile applications make dynamic benchmarks unsuitable as unified evaluation standards. Addressing this challenge is crucial for advancing mobile GUI agent deployment. 
Therefore, we innovatively propose a graph-structured benchmark for complex long-horizon tasks on mobile devices, which bridges the offline-online test discrepancy with enhanced stability compared to dynamic evaluation, addressing the shortcomings of this research. Table~\ref{tab:datasets} compares ColorBench with other mobile GUI agent benchmarks.

\subsection{Comparison with Prior Mobile Agent Graph}
The graph-structured benchmark and the User Interaction Transition Graph (UTG) share a similar high-level structure but differ critically in their granularity and purpose. To meet benchmarking standards, the graph-structured benchmark constructs the topology of state transitions, whereas the UTG defines them based on page relationships. Consequently, the UTGs are primarily used for knowledge exploration during the execution of a single task~\citep{wen2024autodroid,fan2025gui}, representing a subset of the evaluable environment. In contrast, the graph-structured benchmark is designed for evaluating multiple tasks. It aggregates all possible execution paths, including error states, into a comprehensive evaluation environment that serves as a subset of the real world. As such, the benchmark constitutes a superset of the exploration space covered by any single UTG.

Additionally, graphs have extensive application in other aspects of GUI agents. For task generation and evaluation, OmniBench automates the synthesis of complex tasks by combining subtasks based on graphs and utilizes these subtask structure graphs for evaluation~\citep{li2024omnibench}. MobiFlow models tasks as directed acyclic graphs for evaluation purpose\citep{bera2018mobi}. For model training, MobileM3 connects pages collected through breadth-first exploration into graphs to learn UI transition relationships~\citep{wu2024mobilevlm}. Methodologically, Xplore-Agent constructs graph-structured page relationships during exploration~\citep{sun2025gui}, while PG-Agent pre-builts page-relationship graphs of specific applications for RAG to augment knowledge~\citep{chen2025pg}. 
These works demonstrate the advantages of graphs in GUI tasks: precise action transition relationships, natural page navigation modeling, and a clear, controllable global perspective. 
Mobile-Bench-v2 utilizes MobileM3's graphs to generate multi-path tasks, introducing a graph-structured benchmark for the first time, but it lacks systematic explanations~\citep{xu2025mobile}. Our work fills this research gap and proposes a feasible methodology for building a graph-structured benchmark from scratch.


\section{Graph-Structured Benchmark}
\label{sec:graph}
\subsection{Definition of Graph-Structured Benchmark}
The graph abstracts the finite states of real mobile environment into a strongly connected directed graph $G = (V, E)$, comprising two main components:
\begin{itemize}[leftmargin=*]
\item $V=\{N_1[p_1,p_2,\dots],N_2[\dots],\dots\}$: The node set, modeling all screen states $N_i$ that may be encountered in actual evaluations. From the HOME page to detailed app screens and inter-app navigation pages, different screen states are modeled as distinct nodes at the granularity of action transitions. For screens with random elements, 
they are treated as different screens $p_1, p_2,\dots$ within a unified node $N_i$ when recommended content does not involve action transitions. This effectively models the randomness observed in real-device testing.
\item $E=\{(N_i,N_j,a),\dots\}$: The edge set models the action transition relationships between screen states at the node granularity. Each edge corresponds to a specific action a transitioning from one node $N_i$ to another $N_j$, such as ``click bbox[x1,y1,x2,y2]'' or ``type[TEXT]'', aligning one-to-one with the action space used in real-device evaluations.
\end{itemize}
It integrates all tasks into a single, invariant graph structure, simplifying the evaluation pipeline. When evaluating, each task originates from a unified ``HOME'' node. The agent determines actions based on user queries and the current screen, then transitions to the next screen via actions within the graph.

\subsection{Advantages of Graph-Structured Benchmark}  
In this section, we detail the advantages of the graph-structured benchmark over the other two types of evaluation paradigms.
Compared to static evaluation, the graph structure inherently simulates page transitions from a real mobile environment, which allows it to naturally support multi-path solutions to a problem. It is designed to encompass not only the optimal path but also sub-optimal ones and, importantly, recovery paths where the agent corrects its errors. Therefore, graphs offer superior fault tolerance, allowing models to reflect on past errors and return to re-execute without terminating the evaluation. This enables the assessment of collaborative intelligence within agent systems. Compared to dynamic evaluation, the static data construction ensures a stable, controllable, and reproducible testing process. It enables automated evaluation through predefined nodes, which represent task completion, while also avoiding security risks such as unintended payments or information leaks associated with real-world third-party app logins.

In terms of evaluation metrics, this approach breaks through the limitations of static step-level and dynamic result-level assessments. By setting sub-task milestones on the graph, it avoids misjudgments of single-step execution details.
This enables effective evaluation of completion rates for complex, long-horizon tasks composed of multiple atomic tasks, while pinpointing the completion status of subtasks. Consequently, it achieves evaluation at the atomic task capability level, allowing targeted identification of an agent's shortcomings in specific tasks. Furthermore, to simulate real-world environmental variability, multiple images of functionally identical pages are retained. This approach replicates random effects in practical scenarios, striking a balance between the stability of static evaluation and the randomness of dynamic testing. 

\section{ColorBench}

\subsection{Overview and Statistical Analysis}
To construct complex, long-horizon tasks with multiple valid solutions, we employed partially ambiguous instructions, similar to those in Mobile-Bench-v2~\citep{xu2025mobile}, to create a finite and controlled set of correct solutions. Our resulting benchmark, ColorBench, comprises 175 such tasks—101 cross-app and 74 single-app. After excluding random advertisements, pop-ups, and sub-optimal paths, the average optimal path length exceeds 13.13 steps. Representative tasks include price comparison and multi-content sharing (cross-app), as well as food ordering and specific content queries (single-app). The diverse combinations and strong inter-dependencies among subtasks mean each one critically influences the final outcome, thus realistically simulating real-world interactions and narrowing the gap between offline and online evaluation. From the commonalities of these subtasks, we have summarized 15 atomic task capabilities. Figure~\ref{fig:total_statistics} displays the statistics of ColorBench, and the supported action space is detailed in Appendix~\ref{app:action-space}.
\begin{figure}[tb]
    \centering
    \begin{subfigure}[b]{0.26\textwidth}
        \centering
        \includegraphics[width=\textwidth]{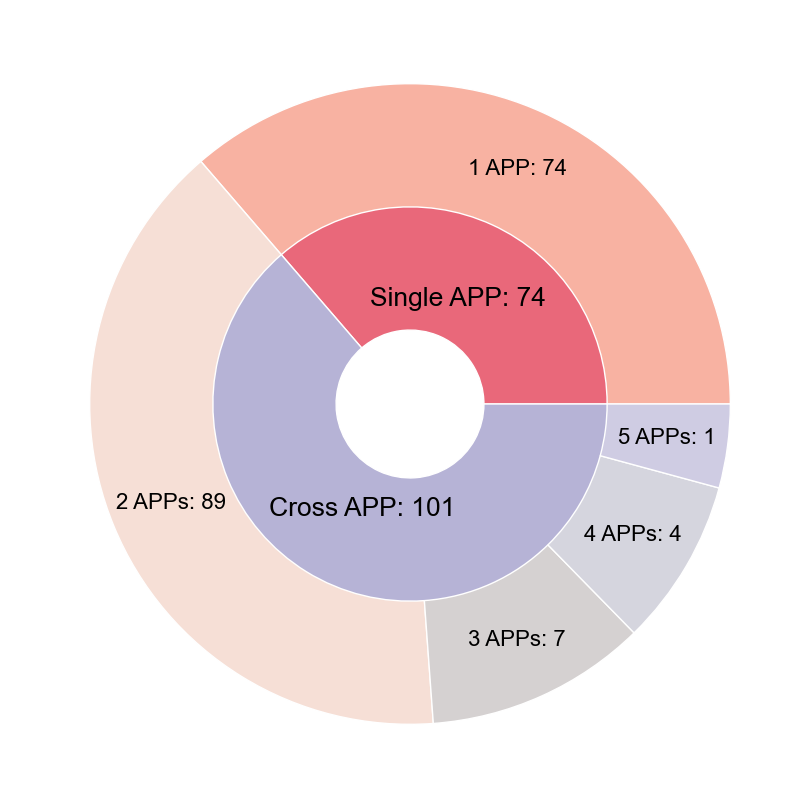}
        \caption{The number of apps included in each task.}
        \vspace{-2mm}
        \label{fig:app_statistics}
    \end{subfigure}
    \hfill
    \begin{subfigure}[b]{0.39\textwidth}
        \centering
        \includegraphics[width=\textwidth]{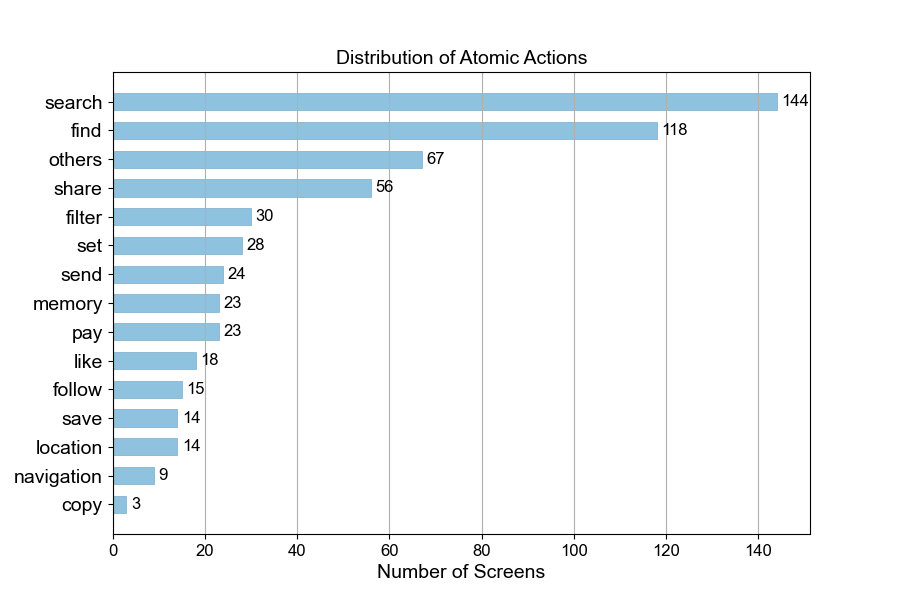}
        \caption{Number of each atomic task used for capability evaluation.}
        \vspace{-2mm}
        \label{fig:atomic_capability_statistics}
    \end{subfigure}
    \hfill
    \begin{subfigure}[b]{0.26\textwidth}
        \centering
        \includegraphics[width=\textwidth]{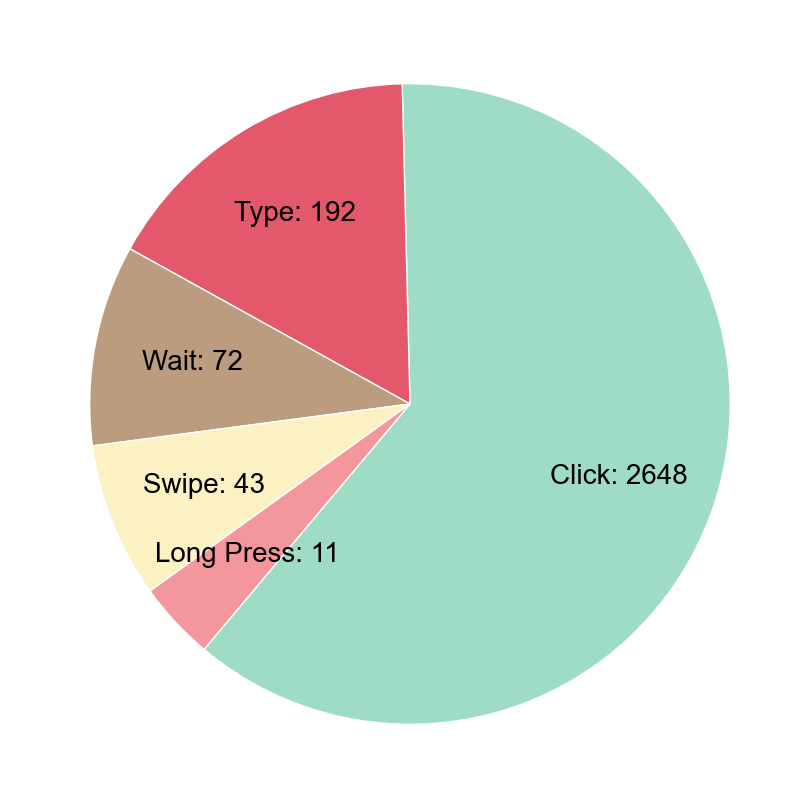}
        \caption{Number of each action in the graph.}
        \vspace{-2mm}
        \label{fig:colorbench_action_statistics}
    \end{subfigure}
    \caption{\textbf{ColorBench statistics.} \textmd{The results do not include actions ``navigate back'', ``navigate home'', or ``open app''.}} %
    \vspace{-4mm}
    \label{fig:total_statistics}
\end{figure}

\subsection{Dataset Construction}
\label{sec:construction-methodology}

We design a graph construction strategy for benchmarking that balances quality and automation. The strategy comprises two primary phases: trajectory collection and graph merging.
The trajectory collection phase aims to fully capture UI elements with high interaction probabilities on key pages within evaluation paths. 
It integrates both breadth-based and depth-based trajectory collection methods, simultaneously covering high-frequency short-distance tasks and complex long-distance tasks. 
The graph merging phase constructs an interactive evaluation environment from the collected high-quality trajectories. This phase employs semantic and visual criteria, utilizing action transitions as the key basis for distinguishing different page nodes, thereby accurately identifying effective interaction boundaries between interface states.
Subsequently, we employ models to automatically annotate complete UI bounding boxes, with the detailed annotation method provided in Appendix~\ref{app:bbox-annotation}. All aforementioned processes underwent final manual quality inspection and refinement. Details regarding the models used in the automation strategy, data collection scales, and cleaning results are introduced in Appendix~\ref{app:graph-construction}. Throughout the construction process, we prioritized the fidelity of the graph in simulating dynamic environments.
The following subsections will elaborate on the complete construction process of ColorBench.
\vspace{-0.2em}
\subsubsection{Trajectory Collection}
\label{sec:trajectory-collection}
\paragraph{Step 1: Breadth-First Search (BFS)}
\label{sec:BFS}
\vspace{-0.5em}
BFS trajectory collection focuses on capturing interactions across shallow-level pages within each application. These pages exhibit high frequency and short-distance characteristics in daily user operations, ensuring the dataset covers the most common trajectories within the target app. Specifically, we employ two VLMs: VLM \textit{A} handles the identification of interactive UI elements. Upon entering a new shallow-level page within the target application, VLM \textit{A} first identifies and stores all interactive UI elements on the page. Subsequently, VLM \textit{B} performs interaction operations on each identified UI element and records the resulting trajectory. This dual-model collaboration ensures accuracy and consistency in the collection process while reducing manual intervention.
\vspace{-0.9em}
\paragraph{Step 2: Depth-First Search (DFS)}
\label{sec:DFS}
DFS trajectory collection targets complex long-horizon tasks that require sustained interaction across multiple applications. Automation tools struggle to capture the complete trajectory of such tasks. To address this, we designed a trajectory collection method based on “screenshot-based action completion” and “branch trajectory supplementation.” This approach effectively reduces the difficulty of collecting complex trajectories while ensuring data integrity and authenticity.

\textbf{Screenshot-Based Action Completion.}
We manually capture screenshots of each step in long-horizon tasks, then use VLM to fill in missing actions between screenshots, thereby reconstructing the complete task trajectory. The process involves three steps: 1) Task Definition and Screenshot Capture: An expert annotation team designs a set of typical, complex long-horizon tasks supporting multiple paths. For each task, trained operators manually execute it on real mobile devices, collecting multiple task completion trajectories and capturing screenshots corresponding to each action. 2) Trajectory Construction: For the acquired sequence of captured screenshots $(S_1, S_2, \dots, S_n)$, each input consecutive screenshots $S_i$ and $S_{i+1}$. The action completion model predicts all correct interaction actions $A_i$. Subsequently, the sequences $(S_1\rightarrow A_1\rightarrow S_2\rightarrow A_2\rightarrow\dots\rightarrow A_{n-1}\rightarrow S_n)$ are combined to form the complete trajectory. 3) Manual Verification: After the model predicts the action, the research team manually verifies all trajectories to ensure their accuracy.

\textbf{Branch Trajectory Supplementation.}
The trajectories obtained through the above steps represent only multiple possible operational paths when humans execute complex long-horizon tasks. Additionally, there may be some correct paths with high model click probabilities (this portion is extremely rare) and erroneous options. These branching paths constitute a crucial component of the model's potential click space, enabling a more realistic simulation of real scenarios. Therefore, we supplement the existing trajectories through the following steps: 1) Let the model perform the same tasks and compare its execution trajectories with the annotated trajectories; 2) Treat other erroneous regions clicked by the model as high-probability potential trajectory spaces, and construct corresponding branch trajectories to supplement the dataset.

\subsubsection{Merge Trajectories into a Graph}
\label{graph-merging}
The graph merging strategy aims to automatically construct evaluation-ready interaction graphs from quality interaction traces. It distinguishes page nodes by identifying action transitions that cause substantive interface state changes. To achieve this goal, we design a two-stage filtering pipeline based on VLMs, with the following workflow:

\textbf{Semantics-Based Coarse Screening.} 
First, we employ large language models to generate semantic descriptions for each screenshot, covering page layout and core functionalities. Subsequently, we compute embedding vectors for all descriptive texts and filter image pairs with semantic similarity exceeding a preset threshold through pairwise comparisons. This step efficiently narrows the candidate pool, focusing on interfaces that are highly similar in both visual appearance and functionality.

\textbf{Action Transition-Based Node Discrimination.}
For image pairs that pass semantic filtering, we feed them into a VLM for fine-grained discrimination. Since candidate pairs are already semantically highly similar, the model's core discrimination criterion shifts from content to whether an ``action transition'' that drives state change exists between them. For instance, interactive operations like ``Click to Follow'' or ``Add to Cart'' may not significantly alter page layouts but trigger substantive updates to interface states, such as button status changes and product count updates. The model classifies image pairs exhibiting such action transitions as belonging to different page nodes, otherwise the same node.

\textbf{Manual Verification and Graph Enhancement.}
After automated merging, we introduced a manual verification process to further enhance the quality of the graph. This step primarily accomplishes two tasks: first, it corrects node classifications that may have been misjudged during automation; second,  it supplements annotations for executable common actions across different state nodes on the same page. Ultimately, we obtained a verified, fully annotated graph that serves as the evaluation environment for ColorBench.

\section{Experiments}

In this section, we conduct extensive experiments to answer the following research questions:
\begin{itemize}
    \item[\textbf{RQ1}] Why graph-structured benchmark are necessary and feasible?
    \item[\textbf{RQ2}] How do existing models perform on complex long-horizon tasks?
    \item[\textbf{RQ3}] What capabilities do existing models lack in complex long-horizon tasks?
    \item[\textbf{RQ4}] Which modules are essential for complex long-horizon tasks?
\end{itemize}

\subsection{Experiment Setup}
\subsubsection{Evaluation Metrics}
To evaluate the performance of models on ColorBench, we leverage SR (success rate) and CR (completion rate)~\citep{liu2025verigui,zhang2024large}, as well as \textbf{Atomic Task Capability (AC)}, which we propose to diagnose capability-level weaknesses that lead to agent failures. In complex long-horizon tasks, each milestone represents the completion of a subtask, and the number of milestones reached reflects the CR of the task. When all milestones are reached, the task is considered successful; otherwise, it is considered a failure.
By extracting common characteristics across all milestone points, we categorize them into 15 atomic task capabilities. For each atomic task, the AC is calculated as:
\begin{equation}
\text{AC} = \frac{\#\text{ Successfully Reached milestones of an Atomic Task}}{\#\text{ the Atomic Task Executed During Evaluation}}.
\end{equation}
The denominator excludes subtasks that were never executed due to failure in preceding subtasks.

\subsubsection{Baselines}
We evaluated Colorbench on open-source models and closed-source models commonly used in mobile GUI agents. Common open-source models for GUI tasks can be categorized into two types: general VLMs such as the Qwen-VL~\citep{bai2023qwenvlversatilevisionlanguagemodel,qwen3vl} series, and specialized foundation models fine-tuned on extensive GUI data based on general VLMs, including UI-TARS~\citep{qin2025ui}, OS-Atlas~\citep{wu2024atlas}, and GUI-OWL series~\citep{ye2025mobile}. Turning to closed-source VLMs, we carefully selected three models: Qwen-VL Max~\citep{qwenvl-max}, GLM-4.5V~\citep{zhipu2024glm45v}, and GPT-4o~\citep{openai2024gpt4osystem}, which are commonly used in mobile GUI agents. Due to the poor grounding capabilities of closed-source models, we employed Qwen2.5-VL-7b~\citep{bai2025qwen2} as the grounding model to identify the target UI element for click and long-press actions.

Moreover, we evaluated two common approaches: 1) we finetuned the base GUI-OWL~\citep{ye2025mobile} model using our static Chinese app training dataset to supplement its knowledge of Chinese applications; 2) we designed a simple multi-agent system incorporating planning, reflection, and memory modules. To investigate the function of these modules for solving complex long-horizon tasks, we conducted ablation experiments on them. Comprehensively considering the model's professional and general ability, we selected Qwen2.5-VL-32B~\citep{bai2025qwen2} and GUI-OWL-32B~\citep{ye2025mobile} for the ablation study. Due to the page limitation, we further provide implementation details in Appendix~\ref{app:experimental-setup}. 
\subsection{Significance of Graph Structure (RQ1)}
\label{sec:feasible-study}
\begin{wrapfigure}{r}{0.5\textwidth}
  \centering
  \vspace{-16pt}
  \includegraphics[width=1.0\linewidth]{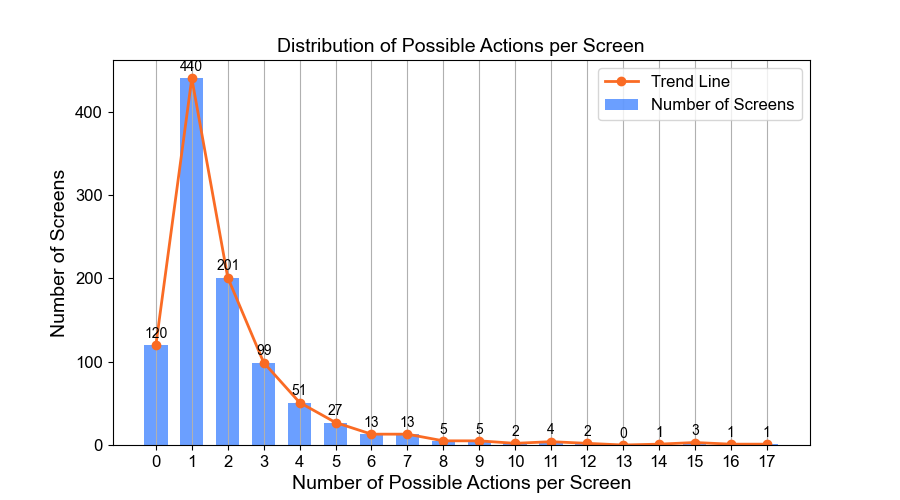}
  \caption{\textbf{Statistics on the number of actions that all experienced models at each node in the graph tend to perform}\textmd{, do not include ``navigate back'', ``navigate home'' and ``open app'' which can be executed on any interface.}}
  \label{fig:preference_action_statistics}
  \vspace{-10pt}
\end{wrapfigure}
The real-world mobile environment is a vast, strongly connected directed graph with an infinite number of nodes and edges, exhibiting immense temporal randomness simultaneously. Replicating such an environment one-to-one is impractical. Given this, a critical question arises: \textit{Can graph-structured evaluation datasets effectively simulate dynamic testing environments to bridge the gap between dynamic and static evaluations?}

To address this question, we conducted two statistical experiments: first, we quantified the potential execution paths for the model operating in a real environment, and second, we compared the outcomes of identical tasks evaluated on a physical device versus on a graph. Specifically, for the first experiment, we analyzed the execution trajectories of all models on ColorBench. By calculating the potential action space at each node (excluding ``navigate back'' and ``home'' actions), we determined the number of successor nodes required at that node in the graph. As shown in Figure~\ref{fig:preference_action_statistics}, the maximum value is 17 corresponding to the APP’s home page, and the minimum value is 0, indicating a task endpoint. The average value is 1.9, reflecting the constructibility of the graph. Consequently, we have demonstrated that the variety of trajectories and screen states that models experienced in real-world evaluations is limited, which validates both the conceptual soundness and practical feasibility of our graph-structured benchmark. 

For the second experiment, we selected a subset of executable tasks from ColorBench for real-device testing, adapting to the dynamically changing real-world information. Partial results are presented in Table~\ref{tab:online-graph} in the appendix. These examples demonstrate that ColorBench covers the primary causes of task failure in real-device evaluations. Therefore, we can assess the authenticity of the graph and demonstrate that a graph-structured benchmark can effectively reduce the gap between offline and online assessments. In fact, when constructing the graph, some of the erroneous trajectories supported by the graph originate from the model’s actual mistakes, which essentially creates a customized testing environment for the model and further ensures the validity of the graph.

\begin{table}[tb]
    \centering
    \fontsize{7.6pt}{8.8pt}\selectfont
    \renewcommand\arraystretch{1.1}
    \caption{\textbf{Performance comparison of closed-source and open-source models on our ColorBench.} \textmd{SR means task success rate and CR means task completion rate (proportion of completed subtasks). Bold represents optimal performance, while underlining represents suboptimal. The results of atomic capabilities are presented in Table~\ref{tab:atomic_capability} in the appendix.}}
    \vspace{-2mm}
    \begin{tabular}{
    >{\raggedright\arraybackslash}p{2.9cm}|
    >{\centering\arraybackslash}p{3.95cm}|
    >{\centering\arraybackslash}p{0.8cm}
    >{\centering\arraybackslash}p{0.8cm}|
    >{\centering\arraybackslash}p{0.8cm}
    >{\centering\arraybackslash}p{0.8cm}|
    >{\centering\arraybackslash}p{0.8cm}
    >{\centering\arraybackslash}p{0.8cm}
    }
    \toprule
    \midrule
    \multirow{2}{*}{\textbf{Baseline}} &\multirow{2}{*}{\textbf{Model}} & \multicolumn{2}{c|}{\textbf{Single APP}}& \multicolumn{2}{c|}{\textbf{Cross APP}} & \multicolumn{2}{c}{\textbf{Average}}\\
    && SR(\%) & CR(\%) & SR(\%) & CR(\%) & SR(\%) & CR(\%)\\
    \midrule
    \multirow{3}{*}{\makecell[l]{\textbf{Closed-source Model} \\ \textbf{with Grounding}}} & GPT-4o & 20.27&	27.59&	11.88&	21.53&	15.43&	24.10 \\
    &Qwen-VL Max & 20.27 &	32.55&14.85&29.42&17.14 &30.74\\
    &GLM-4.5V &  36.49&	56.64&	23.23&	47.72&	28.57&	51.54\\
    \midrule
    \multirow{10}{*}{\textbf{Open-source Model}} &OS-Atlas-Pro-7B & 10.81&	17.91&	3.96&	19.87&	6.86&	19.04\\  
    &UI-TARS-1.5-7B & 9.46&	16.89&	0.00&	13.88&	4.00&	15.15\\  
    &UI-TARS-7B-DPO & 8.11&	12.05&	0.99&	11.82&	4.00&	11.91\\  
    &GUI-OWL-7B& 25.68&	39.19&	16.83&	34.16&	20.57&	36.29\\  
    &Qwen2.5-VL-7B &  22.97&	34.80&	9.90&	32.57&	15.43&	33.51 \\
    &GUI-OWL-32B&36.49&	47.52&	22.77&	39.31&	28.57&	42.78\\  
    &Qwen2.5-VL-32B & 24.32&	36.49&	11.88&	29.62&	17.14&	32.53\\
    &UI-TARS-72B-DPO &33.78&	46.85&	23.76&	45.40&	28.00 &	46.01\\  
    &Qwen2.5-VL-72B & 21.62&	34.23&	16.83&	32.38&	18.86&	33.16\\  
    &Qwen3-VL-235B-A22B-Instruct& 35.14&	49.66&	\textbf{27.72}&	\textbf{43.42}&	30.86&	46.06\\
    \midrule
    \multirow{2}{*}{\makecell[l]{\textbf{Fine-tuned with}\\ \textbf{ColorBench-train}}}&\textbf{GUI-OWL-7B-RL} & 31.08&	46.62&	19.80&	41.79&	24.57&	43.83\\  
    &\textbf{GUI-OWL-32B-RL} & \underline{40.54}&	53.72&	\underline{26.73}&	42.25&	\textbf{32.57}&	\underline{47.10}\\  
    \midrule
    \multirow{2}{*}{\textbf{Multi-Agent System}} &\textbf{GUI-OWL-32B} &\textbf{43.24}&\textbf{55.86}&	22.77&	\underline{43.40} &	\underline{31.43} &	\textbf{48.67}\\
    &\textbf{Qwen2.5-VL-32B} &\underline{40.54}&	\underline{55.32}&	20.79&	40.18&	29.14&	46.58\\
    \midrule
    \bottomrule
    \end{tabular}
    \vspace{-2mm}
    \label{tab:baseline}
\end{table} 

\subsection{Overall Performance (RQ2)}
Table \ref{tab:baseline} presents the main experimental results. For closed-source models, we observe that:
\begin{itemize}[leftmargin=10pt]
    \item GLM-4.5V significantly outperforms both Qwen-VL Max and GPT-4o. Although its SR is not the highest, its notably high CR demonstrates stronger capabilities in comprehending and planning for complex, long-horizon tasks. Furthermore, Table~\ref{tab:atomic_capability} in the appendix indicates that GLM-4.5V can actively memorize essential historical information, an ability that surpasses most compared models. In contrast, GPT-4o underperforms due to a lack of training on relevant operational knowledge, while Qwen-VL Max exhibits deficiencies in decomposing and planning complex tasks. Additionally, all three models demonstrate generally weak UI grounding abilities.
    \item A lack of knowledge about mobile phone operation severely impedes task execution. For instance, despite its powerful multi-modal understanding, GPT-4o's unfamiliarity with basic mobile actions (\ie, copying, pasting, and sharing) often results in it providing only high-level plans without being able to execute the specific operational steps required.
\end{itemize}
As for open-source models, we can draw the following observations from Table~\ref{tab:baseline}:
\begin{itemize}[leftmargin=10pt]
    \item Models with larger parameter scales generally achieve better performance on complex long-horizon tasks. For instance, both the GUI-OWL and Qwen series show improved results as the model size increases, and their largest models, alongside GLM-4.5V, are the only three capable of actively memorizing essential historical information, as shown in Table~\ref{tab:atomic_capability} in the appendix. This stems from their stronger general capabilities, enabling better comprehension and planning for complex tasks.
    \item Specialized foundation models do not necessarily outperform general-purpose models. Fine-tuning can easily lead to overfitting, reducing their ability to generalize to complex long-horizon tasks, as evidenced by the results of UI-TARS and OS-Atlas-Pro (7B). In contrast, the strong performance of the GUI-OWL series indicates that specialized foundation models remain promising and warrant further exploration.
    \item The foundational mobile GUI abilities of these models remain unstable. Although the open-source models have been trained to varying degrees on mobile GUI operations, we still observe issues such as execution step errors, recognition and grounding deviations, and instruction-following failures. Therefore, basic mobile operation capabilities still require improvement. Comparisons with closed-source models further indicate that domain-specific knowledge forms a critical foundation for successfully accomplishing these tasks.
\end{itemize}

As shown in Table~\ref{tab:atomic_capability}, models fine-tuned with app-specific data demonstrate higher task accuracy due to their improved familiarity with application layouts, content, and functionality. In multi-agent systems, decomposing complex tasks into structured modules with enforced execution further enhances performance reliability. Additionally, the appendix records each model’s atomic capability scores, which directly influence the success or failure of the task. Further analysis of these atomic capabilities is provided in Appendix~\ref{app:atomic_capability}.

\subsection{Key Capabilities (RQ3)}
We manually examined the task execution logs of each model and analyzed their performance in essential high-level cognitive capabilities (those requiring reasoning). We identified the following common issues in existing models: incomplete decomposition of complex long-horizon tasks, vague memory of essential historical information, and a lack of effective reflection on recurring erroneous actions.

For example, in the task ``Go to Xiaohongshu to search for `AI Agent', share the paper on `optimized memory management' with WeChat Contact 1, then go to Baidu Arxiv to search and download the PDF of that paper, and finally send it to Contact 1'', some models incorrectly considered the task complete after executing the ``share with contact'' subtask. Other models reached the Arxiv website but failed to retain the specific paper title identified earlier on Xiaohongshu, causing the task to fall into an infinite loop. Most models were unable to recognize their own errors in these situations.

In conclusion, to handle complex long-horizon tasks, models require not only basic GUI capabilities but also the following key advanced competencies: \textbf{the ability to comprehend and analyze complex task requirements, the capacity to decompose long-horizon tasks into structured subtasks, the capability to actively memorize and recall critical information across tasks, and the faculty for reflection and self-correction}. Therefore, the accurate execution of atomic tasks is the foundation for solving complex long-horizon tasks, while high-quality planning, reflection, and memory are the critical support that effectively string these atomic tasks together and ensure their stable execution.

\begin{table}[tbp]
  \centering
  \fontsize{7.6pt}{8.8pt}\selectfont
  \renewcommand\arraystretch{1.1}
  \setlength{\tabcolsep}{2.7pt}
  \caption{\textbf{Ablation experiment results of the multi-agent system modules on our ColorBench.} \textmd{The numbers in the footnote indicate changes relative to the baseline.}}
  \vspace{-2mm}
  \label{tab:system-ablation-table}
  \begin{tabular}{l|ccc|cc|cc|cc}
   \toprule
   \midrule
    \multirow{2}{*}{\textbf{Base Model}} & \multicolumn{3}{c|}{\textbf{Module}} &\multicolumn{2}{c|}{\textbf{Single APP}}& \multicolumn{2}{c|}{\textbf{Cross APP}} & \multicolumn{2}{c}{\textbf{Average}}\\
    & Plan & Reflection & Memory & SR(\%) & CR(\%) & SR(\%) & CR(\%) & SR(\%) & CR(\%)\\
    \midrule
    \multirow{4}{*}{\makecell[l]{\textbf{Qwen2.5-VL}\\\quad\quad\textbf{(32B)}}} & $\times$  & $\times$  & $\times$  & 24.32 & 36.49 & 11.88 & 29.62 & 17.14 & 32.53\\
    & $\surd$  & $\times$ & $\times$  & 32.43\textcolor{green!80!black}{$_{+8.11}$} & 46.28\textcolor{green!80!black}{$_{+9.79}$} & 17.82\textcolor{green!80!black}{$_{+5.94}$} & 35.54\textcolor{green!80!black}{$_{+5.92}$} & 24.00\textcolor{green!80!black}{$_{+6.86}$} & 40.09\textcolor{green!80!black}{$_{+7.56}$}\\
    & $\times$  &$\surd$& $\times$ & 37.84\textcolor{green!80!black}{$_{+13.52}$} & 48.65\textcolor{green!80!black}{$_{+12.16}$} & 15.84\textcolor{green!80!black}{$_{+3.96}$} & 33.48\textcolor{green!80!black}{$_{+3.86}$} & 25.14\textcolor{green!80!black}{$_{+8.00}$} & 39.90\textcolor{green!80!black}{$_{+7.37}$}\\
    & $\times$  & $\times$ & $\surd$ & 31.08\textcolor{green!80!black}{$_{+6.76}$} & 47.52\textcolor{green!80!black}{$_{+11.03}$} & 18.81\textcolor{green!80!black}{$_{+6.93}$} & 35.30\textcolor{green!80!black}{$_{+5.68}$} & 24.00\textcolor{green!80!black}{$_{+6.86}$} & 40.47\textcolor{green!80!black}{$_{+7.94}$}\\
    &$\surd$& $\surd$ & $\times$ & 35.14\textcolor{green!80!black}{$_{+10.82}$} & 51.35\textcolor{green!80!black}{$_{+14.86}$} & 19.80\textcolor{green!80!black}{$_{+7.92}$} & 39.01\textcolor{green!80!black}{$_{+9.39}$} & 26.29\textcolor{green!80!black}{$_{+9.15}$} & 44.23\textcolor{green!80!black}{$_{+11.70}$}\\
    &$\surd$& $\surd$ & $\surd$ & 40.54\textcolor{green!80!black}{$_{+16.22}$} & 55.32\textcolor{green!80!black}{$_{+18.83}$} & 20.79\textcolor{green!80!black}{$_{+8.91}$} & 40.18\textcolor{green!80!black}{$_{+10.56}$} & 29.14\textcolor{green!80!black}{$_{+12.00}$} & 46.58\textcolor{green!80!black}{$_{+14.05}$}\\
    \midrule
    \multirow{4}{*}{\makecell[l]{\textbf{GUI-OWL}\\\quad\quad\textbf{(32B)}}}& $\times$  & $\times$  & $\times$  & 36.49 & 47.52 & 22.77 & 39.31 & 28.57 & 42.78\\
    & $\surd$  & $\times$ & $\times$ & 35.14\textcolor{red}{$_{-1.35}$} & 49.66\textcolor{green!80!black}{$_{+2.14}$} & 22.77\textcolor{gray}{$_{0.00}$} & 46.03\textcolor{green!80!black}{$_{+6.72}$} & 28.00\textcolor{red}{$_{-0.57}$} & 47.56\textcolor{green!80!black}{$_{+4.78}$}\\
    & $\times$  &$\surd$& $\times$ & 37.84\textcolor{green!80!black}{$_{+1.35}$} & 51.35\textcolor{green!80!black}{$_{+3.83}$} & 27.55\textcolor{green!80!black}{$_{+4.78}$} & 46.26\textcolor{green!80!black}{$_{+6.95}$} & 31.43\textcolor{green!80!black}{$_{+2.86}$} & 48.45\textcolor{green!80!black}{$_{+5.67}$}\\
    & $\times$  & $\times$ & $\surd$ & 36.49\textcolor{gray}{$_{0.00}$} & 49.55\textcolor{green!80!black}{$_{+2.03}$} & 25.74\textcolor{green!80!black}{$_{+2.97}$} & 43.86\textcolor{green!80!black}{$_{+4.55}$} & 30.29\textcolor{green!80!black}{$_{+1.72}$} & 46.27\textcolor{green!80!black}{$_{+3.49}$}\\
    &$\surd$& $\surd$ & $\times$ & 41.89\textcolor{green!80!black}{$_{+5.40}$} & 51.91\textcolor{green!80!black}{$_{+4.39}$} & 22.77\textcolor{gray}{$_{0.00}$} & 44.82\textcolor{green!80!black}{$_{+5.51}$} & 30.86\textcolor{green!80!black}{$_{+2.29}$} & 47.82\textcolor{green!80!black}{$_{+5.04}$}\\
    &$\surd$& $\surd$ & $\surd$ &43.24\textcolor{green!80!black}{$_{+6.75}$} & 55.86\textcolor{green!80!black}{$_{+8.34}$} & 22.77\textcolor{gray}{$_{0.00}$} & 43.40\textcolor{green!80!black}{$_{+4.09}$} & 31.43\textcolor{green!80!black}{$_{+2.86}$} & 48.67\textcolor{green!80!black}{$_{+5.89}$}\\
   \midrule
   \bottomrule
   \end{tabular}
   \vspace{-2mm}
\end{table}

\subsection{Ablation Study (RQ4)}

Ablation studies on individual modules reveal their distinct contributions to solving complex long-horizon tasks. Table~\ref{tab:system-ablation-table}, while each module improves overall performance, the extent of this improvement varies across models. For instance, introducing only the reflection module brings substantial gains for Qwen-2.5-VL-32B but limited improvement for GUI-Owl-32B, reflecting differences in their inherent capabilities.

We also observed nuanced interactions between modules. When Qwen-2.5-VL-32B evolves from using only reflection to incorporating both reflection and planning, SR on single-app tasks decreases. Meanwhile, CR increases and remains higher than when planning alone is used. Conversely, GUI-Owl-32B’s CR on cross-app tasks declines as more modules are added, though it still exceeds the baseline. This highlights instability introduced by multi-agent systems. Overly complex module combinations can cause erroneous coupling and error accumulation. As a result, tasks that were initially solvable may become unresolvable.

We attribute this to imbalanced capability distributions, where weaker modules can inhibit stronger ones. This occurs because increased low-quality information raises systemic entropy, while high-quality signals are underutilized. Therefore, multi-agent approaches to long-horizon tasks require not only individual module effectiveness but also balanced integration—otherwise, they risk degrading performance.
\vspace{-1em}

\section{Conclusion}

To bridge the gap between offline static evaluation and online dynamic evaluation for Mobile GUI Agents, we propose a graph-structured benchmark with an effective construction methodology. Based on this framework, we develop ColorBench, a benchmark specifically designed for complex long-horizon tasks that balances static stability and dynamic randomness through finite-state modeling of a dynamic environment. ColorBench supports multiple valid solutions for individual tasks and extends evaluation from step-level and result-level metrics to atomic task capability assessment, enabling effective diagnosis of model deficiencies. We extensively evaluate ColorBench across numerous models, validating the necessity and feasibility of the graph-based benchmark. Based on experimental results, we analyze limitations of existing models and provide concrete suggestions and potential approaches to enhance agents' capabilities in solving complex long-horizon problems.

\bibliography{main}
\bibliographystyle{rlc}

\appendix

\section{Additional Information of ColorBench}
\subsection{Action Space}
\label{app:action-space}
Our ColorBench includes nine actions, and during evaluation, the outputs of different models need to be aligned to the action space.

\begin{table}[htbp]
  \centering
  \caption{Action space of ColorBench.}
  \vspace{-0.2cm}
  \label{tab:action_space}
  \begin{tabular}{l|c}
    \toprule
    \midrule
    Action Type & Parameter\\
    \midrule
    click & coordinate$=(x,y)$ \\
    long press& coordinate$=(x,y)$ \\
    swipe & direction$\in \{\text{up},\text{down},\text{left},\text{right}\}$\\
    type & content=[TEXT] \\
    wait & coordinate$=(x,y)$ \\
    open & app$=\text{app name}$ \\
    navigate back & none \\
    navigate home & none \\
    complete & answer=[TEXT] \\
    \midrule
    \bottomrule
  \end{tabular}
  \vspace{-0.2cm}
\end{table}

\subsection{Annotation of Bounding Box}
\label{app:bbox-annotation}
Accurate bounding boxes are crucial for GUIs, as they define the precise spatial scope of interactive UI elements, preventing mapping errors between user actions and irrelevant areas. Therefore, this paper proposes an annotation method combining multiple VLM integration with human verification, divided into three concise steps:
\begin{itemize}
    \item Step 1: Input actual interaction point coordinates and corresponding images into two VLMs, each generating an interaction region. One VLM outputs a larger bounding box (ensuring complete target coverage), while the other outputs a smaller bounding box (minimizing background inclusion).
    \item Step 2: A third VLM evaluates the two candidate boxes and selects the optimal one. If neither candidate is satisfactory, a new bounding box is generated.
    \item Step 3: Domain experts verify the preliminary results from Step 2, correcting errors (e.g., missing target parts, redundant background) and confirming the final annotation results.
\end{itemize}

\subsection{Supplement on ColorBench Construction}
\label{app:graph-construction}

\subsubsection{Automated Strategy Model Selection}

Our automated pipeline leverages specialized models according to their core capabilities. Qwen2.5-VL-72B, chosen for its strong general reasoning, handles UI element identification and interaction during broad-coverage trajectory collection, predicts the correct action between model outputs for deep trajectory action annotation, and generates page state descriptions with visual similarity judgment during graph merging. Page descriptions are encoded by the ``models--BAAI--bge-large-zh-v1.5'' model for semantic similarity calculation. GUI-OWL-32B, recognized for its exceptional visual grounding capability, is employed exclusively for bounding box annotation.

\subsubsection{Data Collection and Quality Assurance Results}
During the BFS phase, trajectory collection was performed for four individual applications: Weixin, Meituan, Jingdong, Xiaohongshu, yielding 6300 trajectory screenshots. During the DFS phase, an additional 1343 trajectory screenshots for complex long-horizon tasks (including single-app and cross-app ) were collected. Using a graph merging strategy to integrate these trajectories into a unified graph structure, we conducted manual quality control to remove near-duplicate images from identical nodes, retaining only a limited set of representative screenshots. To address coverage gaps in the graph, we manually supplemented 50 screenshots. The final constructed graph contains 1989 carefully curated screenshots.

\subsubsection{System Prompt in Graph Merging}
Figure~\ref{fig:prompt} presents the system prompt template used in our automated graph merging process. Due to page limitations, additional implementation details will be made available in our future open-source release.
\begin{figure}[ht]
\begin{response}[Generating Page Description]
You are a GUI AGENT. Please define in one sentence what page the given mobile screenshot represents, and provide a general description of the page layout, its purpose, and other key information. You should ignore changes in the screenshot caused by time or updates, and ultimately form a brief textual description.
\end{response}
\vspace{2mm}
\begin{response}[Judgement of Same Screen State]
You are a GUI AGENT. Please ignore changes in the screenshots caused by time, and judge whether the two given images belong to the same page state based on aspects such as page layout, page function, and action relationships. \\

Notes: 
1. Treat all mobile home screen pages as the same page;\\
2. Ignore changes in the page caused by time and updates to recommended content, and focus only on changes caused by actions;\\
3. Pay special attention to navigation bar tabs. In the application, different navigation bar tabs indicate different pages;\\
4. Page state changes caused by actions should be considered different pages. If performing an action on one image is required to turn it into the other image, then these two images represent different page states.
\end{response}
\caption{
\textbf{System prompt templates for merging graph.} \textmd{In practical use, given the language of the application, we employed Chinese-language prompts.}
}
\label{fig:prompt}
\end{figure}

\section{Experimental Supplement}
\subsection{Implementation Details}
\label{app:experimental-setup}
All open-source models tested in our experiments were deployed using VLLM(0.10.1) on A100 80G hardware. Since each node in the image dataset may contain multiple images, we implemented random returns and set the random seed to 2025. We configured model inference parameters as follows: temperature=0.1, top\_k=5, top\_p=0.9, max\_tokens=1024. To ensure experimental fairness, we uniformly employed the open-source prompts provided by each model while preserving their original action spaces and performing additional alignment on outputs. 
We also designed identical prompts for closed-source models. For these, we supplemented each input with the model's historical output records and captured its current reasoning and actions.
We will open-source the project code in the future.

\subsection{Result of Atomic Capabilities}
\label{app:atomic_capability}
Table~\ref{tab:atomic_capability} presents evaluation results for atomic task capabilities across open-source and closed-source models. A ``$-$'' in the table indicates the model never executed that subtask, having failed at a preceding subtask, thus rendering the subtask capability unverifiable. 
Higher scores in the table do not directly reflect the model's capability; rather, it is the poorer results that effectively reveal the causes of task failure. For example, GLM-4.5V exhibits high SR and CR but a ``save'' score of 0, indicating it cannot perform the save action. UI-TARS-72B-DPO also exhibits high SR and CR, yet its ``filter'' score is as low as 22.22, indicating this is the critical factor constraining the model's success.

\begin{table}[htb]
  \centering
  \caption{\textbf{Success rate of atomic capabilities of closed-source and open-source models on our ColorBench.} \textmd{All values are in percentage (\%). The symbol `$-$' indicates that the model did not encounter this type of subtask during the evaluation process, usually due to the failure of a preceding task.}}
  \label{tab:atomic_capability}
  \adjustbox{max width=\linewidth}{%
  \begin{tabular}{l|ccccccccccccccc}
   \toprule
   \midrule
    \textbf{Model} & \textbf{follow}& \textbf{pay} &\textbf{save}&\textbf{search}&\textbf{share}&\textbf{set}&\textbf{find}&\textbf{copy}&\textbf{filter}&\textbf{like}&\textbf{send}&\textbf{location}&\textbf{navigation}&\textbf{others}&\textbf{memory}\\
    \midrule
    GPT-4o& 100.00&	50.00&	50.00&	33.06&	76.47&	84.62&	56.67&	0.00&	18.75&	80.00&	25.00&	25.00&	0.00&62.50&0.00\\
    Qwen-VL Max& 40&33.33&	16.67&	43.8&	70.37&	83.33&	60.56&	33.33&	36.36&	80&	50.00&	25.00&	- &68.57&	0.00\\
    GLM-4.5V &90.91&55.56&0.00&72.95&90.62&75.00&76.34&	50.00&26.32&81.82&77.78&66.67&75.00&60.00&45.45\\
    \midrule
    OS-Atlas-Pro-7B & 100.00&	0.00&	75.00&	43.22&	35.71&	66.67&	35.94&	0.00&0.00&71.43&25.00&40.00&100.00&	26.09&0.00\\ 
    UI-TARS-1.5-7B& 50.00&	0.00&	33.33&	30.25&	7.69&	40.00&	37.93&	100.00&	20.00&	100.00&0.00&20.00&0.00&	40.00&0.00\\  
    UI-TARS-7B-DPO& 50.00&	0.00&	50.00&	24.58&	37.50&	66.67&	25.49&	0.00&	22.22&	71.43&	0.00&	0.00&-&34.78&0.00\\  
    GUI-OWL-7B&85.71&60.00&25.00&57.85&	88.46&	69.23&	62.67&	50.00&11.11&75.00&50.00&33.33&0.00&	58.97&0.00\\  
    Qwen2.5-VL-7B&  83.33&	75&	16.67&	61.48&	50.00&	75.00&	52.00&	100.00&25&	77.78&	83.33&	16.67&	0&	51.52&0.00\\
    GUI-OWL-32B &87.50&	62.50&0.00&	64.23&	81.82&	84.62&	64.10&50.00&31.25&	100.00&	57.14&	40.00&	66.67&	69.23&	0.00\\  
    Qwen2.5-VL-32B & 33.33&	83.33&	0.00&	54.55&	71.43&	84.62&	52.05&	50.00&	16.67&	88.89&	75.00&	50.00&	50.00&	59.38&0.00\\
    UI-TARS-72B-DPO & 77.78&	55.56&	28.57&	68.50&	85.00&	85.71&	63.29&	100.00&	22.22&	88.89&	66.67&	50.00&	66.67&	73.17&	28.57\\  
    Qwen2.5-VL-72B& 42.86&	75.00&	0.00&	61.48&	66.67&	84.62&	52.00&	50.00&	6.67&	90.00&	60.00&	20.00&	50.00&	54.84&0.00\\  
    Qwen3-VL-235B-A22B-Instruct & 90.00&	71.43&	28.57&	62.90&	86.49&	92.31&	68.75&	100.00&35.00&	90.91&	83.33&	16.67&0.00&71.43&50.00\\
    \midrule
   \bottomrule
   \end{tabular}}
\end{table}

\begin{table}[t!]
  \centering
  \caption{\textbf{Comparison of some failure results evaluated using the ColorBench task on real devices and on the graph dataset.} \textmd{Failure reasons are recorded separately for the specific reasons under the two evaluation environments. The tasks in the table are actually evaluated in Chinese.}}
  \label{tab:online-graph}
  \fontsize{7.6pt}{8.8pt}\selectfont    
  \renewcommand\arraystretch{1.1}    
  \setlength{\tabcolsep}{2.7pt}
  \begin{tabularx}{\linewidth}{>{\RaggedRight\arraybackslash}p{0.4\linewidth} >{\RaggedRight\arraybackslash}p{0.31\linewidth} >{\RaggedRight\arraybackslash\sloppy}p{0.35   \linewidth}} %
    \toprule
    \midrule
    \textbf{Task} & \textbf{Failure Reason of Real-Device} & \textbf{Failure Reason of Graph}\\
    \midrule
    Search for ``Guangzhou travel guides'' on Xiaohongshu, share the first one with Xiaohongshu friend Hh Yuan, and ask ``Which place do you want to visit the most?'' Then use Xiaohongshu to search for food near Canton Tower, remember the name of the first restaurant, search for that restaurant on Meituan and order a set meal for two. & Accidentally clicked on the advertisement when searching for the store on Meituan.	& Failed to search for food on \quad\quad\quad Meituan.\\
    \midrule
    Search for ``Guangzhou travel guides'' on Xiaohongshu, find the first one, check the blogger's followers, likes, and favorites, and then tell WeChat friend 1. & There is an error in the data sent to WeChat friends. & Did not enter the blogger's \quad\quad\quad\quad homepage.\\
    \midrule
    Search for ``Guangzhou travel guides'' on Xiaohongshu, use the ``Q\&A'' to let AI reply, and save the generated content as an image to the album.& Fail to find ``Q\&A''. &Failed to save as an image.\\
    \midrule
    Compare the price of the ``Xin Dou Ji Guangzhou Tower'' double set meal on Meituan, Dazhong Dianping, and TikTok, then forward the cheapest one to friend 1. &	It ends after executing on Meituan.&Failed to search after Meituan.\\
    \midrule
    Search for ``Baheli Beef Hotpot'' on Meituan, forward all double set meals to WeChat friend 1, and ask him to ``choose one'' after sharing.& 0	No sharing, just repeatedly clicking to grab the deal. & Only shared one set meal.\\
    \midrule
    Order takeaway from ``Baheli Beef Hotpot'' on Meituan: one dry-fried beef, without dinnerware. & Mistakenly viewed the in-store set meal instead of selecting takeout. &	Did not accurately find takeout.\\
    \midrule
    Search for ``Baheli Beef Hotpot'' on Meituan, choose the first restaurant and navigate directly using Gaode Map.& Did not use Gaode Map navigation.&Did not switch to Gaode Map. \\
    \midrule
    Go to JD xxx official flagship store, search for ``a certain mobile phone'' in the store, check what colors are available for the 16+512GB configuration, and then tell WeChat friend 1. &Did not enter the official flagship store.&Did not search in the official \quad\quad\quad\quad\quad flagship store.\\
    \midrule
    Go to JD xxx official flagship store, forward the store detail page to Moments.& Could not find the store details page, only accessed the homepage.&Could not find the store details \quad\quad\quad page.\\
    \midrule
    Go to JD xxx official flagship store, share the homepage with WeChat friend 1.&  Could not find the share UI. &	Could not find the share option.\\
    \midrule
    \bottomrule
  \end{tabularx}
\end{table}

\end{document}